\documentclass[10pt, a4paper]{article}

\usepackage{lrec2026} 
\usepackage{linguex}     
\usepackage{cgloss4e}    
\usepackage{tipa}        
\usepackage{amsmath,amssymb}

\newcommand{\gl}[1]{\textsc{#1}}
\usepackage{xcolor} 
\usepackage{tikz}
\usetikzlibrary{arrows.meta,positioning,shapes.misc,calc}

\title{Plausibility as Commonsense Reasoning: Humans Succeed, Large Language Models Do not}

\name{Sercan Karaka{\c{s}}}

\address{University of Chicago \\
         skarakas@uchicago.edu\\}

\abstract{
Large language models achieve strong performance on many language tasks, yet it remains unclear whether they integrate world knowledge with syntactic structure in a human-like, structure-sensitive way during ambiguity resolution. We test this question in Turkish prenominal relative-clause attachment ambiguities, where the same surface string permits high attachment (HA) or low attachment (LA). We construct ambiguous items that keep the syntactic configuration fixed and ensure both parses remain pragmatically possible, while graded event plausibility selectively favors High Attachment vs.\ Low Attachment. The contrasts are validated with independent norming ratings. In a speeded forced-choice comprehension experiment, humans show a large, correctly directed plausibility effect. We then evaluate Turkish and multilingual LLMs in a parallel preference-based setup that compares matched HA/LA continuations via mean per-token log-probability. Across models, plausibility-driven shifts are weak, unstable, or reversed. The results suggest that, in the tested models, plausibility information does not guide attachment preferences as reliably as it does in human judgments, and they highlight Turkish RC attachment as a useful cross-linguistic diagnostic beyond broad benchmarks.
\\ \newline 
}

\begin{document}

\maketitleabstract

\thispagestyle{empty}
\pagestyle{empty}
\pagenumbering{gobble}

\section{Introduction}

Recent progress in large language models (LLMs) has produced systems that reach state-of-the-art results on a wide range of language tasks, including open-domain question answering, generation, summarization, creative writing and translation (\citealp{anil2023palm2,openai2024gpt4technicalreport,riviere2024gemma2,li2025appleintelligencefoundationlanguage,singh2025openaigpt5card,gemmateam2025gemma3technicalreport,finkelstein2026translategemmatechnicalreport}). In addition to this broad linguistic competence, LLMs often exhibit strong performance in demanding problem-solving workloads, such as software development and symbolic mathematics. In parallel, psycholinguistics and cognitively oriented computational linguistics have increasingly treated neural language models as explicit probabilistic theories of incremental comprehension. Surprisal-based accounts have long been used to explain processing difficulty in reading-time data and related measures. Foundational work showed that surprisal derived from symbolic or probabilistic grammars predicts human comprehension difficulty, including effects related to syntactic ambiguity resolution and expectation-based parsing (\citealp{hale2001,levy2008}); more recent work extends this approach to surprisal derived from neural language models (\citealp{arehalli-etal-2022-syntactic}). At the same time, a foundational issue remains open: whether these models carry out human-like reasoning that is logically compositional and systematically generalizes, or whether their apparent success primarily reflects the ability to exploit massive training data to reproduce high-probability surface regularities. Several diagnostic evaluations suggest that LLM behavior often reflects probabilistic pattern completion or shallow heuristics rather than fully systematic rule-based inference \citep{mccoy-etal-2019-right,yang-etal-2024-exploring}. This appears to be especially visible under controlled compositional generalization settings where models often struggle to recombine familiar primitives in genuinely novel ways (\citealp{lake2018generalization,ruis2020gscan}).

Furthermore, interest in using large language models as cognitive models of human language processing has grown rapidly in recent years, but the empirical foundation remains strongly English-centric: comparatively few studies have carried out systematic, theory-driven psycholinguistic evaluations in other languages, limiting the generalizability of current conclusions across typologically diverse settings (\citealp{futrell-etal-2019-neural,joshi-etal-2020-state,hollenstein-etal-2021-multilingual,alves-2025-benchmarking,boeve-bogaerts-2025-dutch}). Accordingly, we study Turkish, which is a morphologically rich language that is nonetheless less represented in widely used NLP training resources and evaluation benchmarks (\citealp{conneau-etal-2020-unsupervised,joshi-etal-2020-state}), and focus on a particularly informative phenomenon, which is relative-clause attachment ambiguity. In Turkish prenominal relative clause (RC) configurations, the clause can in principle modify either the higher noun (high attachment; HA) or the lower noun (low attachment; LA), yielding a classic hierarchical ambiguity under a fixed surface string. In the broader sentence-processing literature, ambiguity resolution in high- versus low-attachment configurations has often been characterized in terms of economy-driven, rightward-attachment preferences---Late Closure \citep{frazierfodor1978sausagemachine}, Right Association \citep{kimball1973seven}, and Recency \citep{gibson1996recency}---while cross-linguistic work has tested the scope and limits of these biases \citep{baccino2000lateclosure,fernandez2003bilingual}. At the same time, a large body of evidence shows that attachment preferences are not determined by structural heuristics alone: discourse and semantic context can substantially modulate parsing commitments, including in reduced relative-clause ambiguities \citep{spivey1993context}.

Building on this line of work, we focus on a more fine-grained cue than broad pragmatic \emph{possibility}. Specifically, we isolate graded \emph{world-knowledge plausibility}: we hold the syntactic environment constant while selectively biasing the plausibility of attachment to the higher vs.\ lower nominal, and we validate these contrasts via independent norming ratings. We then test whether such plausibility guides attachment in Turkish by combining (i) an online speeded forced-choice comprehension experiment with native Turkish speakers and (ii) a parallel evaluation of three LLMs using matched log-probability scoring. The two settings are aligned at the level of the underlying interpretive contrast, not at the level of task mechanics: humans make an explicit attachment choice after reading the sentence, whereas models are evaluated through their relative preference for matched HA versus LA continuations. Accordingly, we treat the human response and the model log-probability preference as parallel but non-identical proxies for attachment preference under the same plausibility manipulation. Our results demonstrate a robust human sensitivity to plausibility: attachment preferences shift substantially in the predicted direction when world-knowledge cues favor high attachment (HA) versus low attachment (LA). In contrast, the LLMs show markedly weaker and less consistent plausibility-based shifts, with their preferences often seeming to reflect structural attachment biases rather than the intended plausibility manipulation.

The rest of the paper is structured as follows. \S\ref{sec:related} reviews related work, and \S\ref{sec:background} introduces Turkish relative clause attachment. \S\ref{sec:experiments} presents the human experiment, while \S\ref{sec:models} outlines the LLM evaluation setup. We then present the human and model results, including the model-specific findings in \S\ref{sec:model-results}, before discussing their implications for world-knowledge integration and the use of LLMs as cognitive models in \S\ref{sec:discussion}.

\section{Related Work}
\label{sec:related}

\subsection{Attachment ambiguity and constraint interaction}
Relative-clause attachment ambiguities have long served as a central testbed for theories of incremental parsing, because a single surface string can license multiple hierarchical structures. Early accounts emphasized economy- and locality-based heuristics such as Late Closure \citep{frazierfodor1978sausagemachine}, Right Association \citep{kimball1973seven}, and Recency \citep{gibson1996recency}, which predict a general bias toward attaching incoming material to the most recently processed constituent. A broad cross-linguistic literature has evaluated the scope of these principles and documented substantial variability, motivating proposals that attachment is not governed by a single universal heuristic, but is instead sensitive to language-specific distributions and to interactions among multiple sources of information \citep{baccino2000lateclosure,fernandez2003bilingual}.

In parallel, constraint-based approaches argue that syntactic commitments are continuously shaped by probabilistic cues from semantics, discourse, and world knowledge, often well before disambiguating material arrives. Classic demonstrations show that supportive discourse and semantic context can dramatically reduce garden-path difficulty in reduced RCs \citep{spivey1993context,altmannsteedman1988context}. Related work on ``thematic fit'' further indicates that event knowledge and plausibility can be quantified independently (e.g., via norming) and can guide online interpretation early in processing \citep{mcrae1998thematicfit}.

\subsection{Prenominal RC attachment in Turkish}
Turkish provides a particularly informative setting for attachment research because RCs are typically prenominal and because attachment ambiguities often arise in configurations with multiple potential hosts inside complex nominal structures. Early work highlighted that Turkish had been comparatively less explored in the RC-attachment literature and reported that attachment outcomes are strongly modulated by lexical-semantic properties of the potential hosts (e.g., animacy, semantic compatibility) and by properties of the complex NP environment, challenging accounts that rely on syntactic locality alone \citep{kirkici2004turkishrc}. Subsequent studies likewise emphasize the importance of controlling semantic relations within the nominal complex; when such factors are carefully balanced, Turkish comprehenders may show no stable baseline preference for one attachment site over the other, suggesting that apparent ``preferences'' can reflect uncontrolled semantic biases rather than a fixed parsing strategy \citep{baser2020turkishrc}.

Other work has asked more directly whether locality-style biases surface under particular timing or structural conditions. For example, experiments manipulating predicate proximity and related locality factors report patterns consistent with recency-driven attachment in at least some Turkish materials, while also indicating that additional structural factors can shift rates of high vs.\ low attachment \citep{akal2021recency}. More recent online studies connect Turkish prenominal RC attachment to broader debates about ambiguity advantage and underspecification: under some accounts, readers may delay commitment to save effort, whereas race-based models predict different processing signatures for prenominal structures. Evidence from Turkish reading experiments has been used to argue against strong underspecification-based predictions in this domain, and plausibility has been employed as a controlled disambiguating cue in constructing attachment conditions \citep{logacev2022underspecification}.

Our contribution, on the other hand, is to move beyond relatively coarse manipulations of pragmatic \emph{possibility} and instead isolate a narrower, graded world-knowledge \emph{plausibility} cue. We construct minimally contrasting Turkish materials in which the syntactic configuration is held constant while the plausibility of attachment is selectively biased toward the higher vs.\ lower nominal, and we validate these contrasts with independent norming ratings (cf.\ thematic-fit methodologies; \citealp{mcrae1998thematicfit}). This lets us ask whether attachment decisions track the intended plausibility gradient, rather than reflecting uncontrolled lexical associations.

\subsection{Neural language models as cognitive models, and evaluation beyond English}
A growing body of psycholinguistic and cognitively oriented computational work uses neural language models as probabilistic models of incremental comprehension, asking to what extent their surprisal estimates and internal representations align with human sentence-processing behavior \citep{futrell-etal-2019-neural,oh-schuler-2023-transformer,arehalli-etal-2022-syntactic}. As mentioned before, in this view, surprisal derived from a model’s next-word distribution provides a linking hypothesis to processing difficulty, with influential proposals and evidence showing that surprisal predicts reading-time patterns and ambiguity-resolution effects in humans \citep{hale2001,levy2008}. Recent work extends these ideas to modern neural architectures and explores syntactic ambiguity phenomena under LM-based predictors \citep{arehalli-etal-2022-syntactic}. At the same time, targeted ``psycholinguistic diagnostics'' highlight systematic mismatches between model behavior and human generalizations, especially for inferences that require robust integration of context, roles, or negation \citep{ettinger2020bert}. Related lines of work further investigate when distributional models do (or do not) acquire pragmatic constraints that humans exploit in disambiguation \citep{davis2020rnncontext}.

Importantly for the present paper, much LM-as-cognition evidence remains concentrated on English, and cross-linguistic generalizations are therefore less secure. This motivates testing typologically different languages to evaluate whether purportedly general cognitive conclusions about LM surprisal, ambiguity resolution, and cue integration persist under different structural priors and training-resource profiles. While recent work has begun to extend LM-as-cognition evaluations to Turkish---both via human--LLM comparative processing studies and via cross-linguistic surprisal validations that include Turkish---the overall evidence base remains comparatively sparse \citep{de-varda-marelli-2022-effects,keles-deniz-2024-superficial,karakas2026clauseinternalclauseexternaltestingturkish}. We aim to leverage this gap by comparing human RC attachment judgments and model attachment preferences under the same controlled, normed plausibility manipulation in Turkish. Our goal is not to equate the human task with the model evaluation procedure, but to test whether the same item-level plausibility contrast shifts attachment in a consistent direction across the two systems, or whether model behavior is instead dominated by surface regularities and base-rate biases.

\section{Turkish relative clauses}
\label{sec:background}

Turkish relative clauses are typically \emph{prenominal}: the RC precedes the noun it modifies, and relativization is expressed morphologically on the verb, commonly via the nominalizer/participle in RCs. In complex noun phrases with two potential nominal hosts, the same prenominal RC string can in principle modify either the higher noun (high attachment; HA) or the lower noun (low attachment; LA), yielding a hierarchical ambiguity under an otherwise fixed surface string. 

\ex.\label{ex:kirkici-balcony}
\gll Birileri, balkon-da dur-an [aktris-in]\textsubscript{\footnotesize LOW}
     [hizmetçi-si-ni]\textsubscript{\footnotesize HIGH} vur-du. \\
     someone-\gl{nom} balcony-\gl{loc} stand-\gl{rel}
     actress-\gl{gen} servant-\gl{poss}-\gl{acc} shoot-\gl{pst} \\
\glt `Someone shot the servant of the actress who was on the balcony.'
\hfill (\citealt{kirkici2004turkishrc}: 4--5)

In \ref{ex:kirkici-balcony}, the prenominal relative clause balkon-da dur-an `standing on the balcony' can modify either noun inside the complex NP. Under low attachment, the RC modifies the lower noun \emph{aktris} `actress', yielding the reading `Someone shot the servant of the actress who was on the balcony.'
' Under high attachment, the RC modifies the higher noun hizmetçi `servant', yielding the reading `Someone shot the servant (of the actress) who was on the balcony.' Because both hosts are structurally available, the surface string remains fixed while the attachment site determines the intended referent of the RC.

\section{Experiments}
\label{sec:experiments}

We recruited $102$ native speakers of Turkish. To ensure data quality, we excluded
$16$ participants whose response latencies were extremely fast or extremely slow
relative to the sample distribution. Specifically, for each participant $i$, we
first computed their mean response time across trials:

\newcommand{\formexgap}{\vspace{0.35\baselineskip}}

\ex.\label{ex:rt-participant-mean}
\[
\bar{t}_i = \frac{1}{N_i}\sum_{j=1}^{N_i} t_{ij}
\]
\noindent where $t_{ij}$ is the response time on trial $j$ and $N_i$ is the number
of trials completed by participant $i$.

\formexgap

We then computed the grand mean and standard deviation across participants:

\ex.\label{ex:rt-grand-mean-sd}
\[
\mu = \frac{1}{P}\sum_{i=1}^{P}\bar{t}_i,
\qquad
\sigma = \sqrt{\frac{1}{P-1}\sum_{i=1}^{P}\left(\bar{t}_i-\mu\right)^2}
\]

\formexgap

Finally, we excluded participant $i$ if their mean response time satisfied the
following outlier criterion:

\ex.\label{ex:rt-threshold}
\[
\bar{t}_i < \mu - 2\sigma
\qquad \text{or} \qquad
\bar{t}_i > \mu + 2\sigma
\]

After these exclusions, $86$ native speakers of Turkish remained for the online experiment, which was administered using the \textsc{PCIbex}/\textsc{PennController} platform \citep{zehrschwarz2018penncontroller}.

\paragraph{Materials and design.}

Previous work on Turkish (\citealt[251]{baser2018syntactic}) uses examples such as \ref{ex:baser-school} to illustrate \emph{relative-clause attachment ambiguity} inside a complex NP. In this configuration, the prenominal RC precedes two potential nominal hosts, the lower noun and the higher noun. Structurally, the RC can in principle modify either NP1 or NP2, so the surface string is compatible with two parses. Crucially, however, the contrast in this type of item is not a subtle plausibility manipulation. In \ref{ex:baser-school}, the RC \emph{okula kaydedilen} `who was enrolled in the school' can in principle modify either \emph{müdür} `principal' or \emph{yeğen} `nephew', but only the latter yields a semantically coherent interpretation. Attaching the RC to \emph{müdür} is structurally available, yet semantically anomalous, since a principal is not normally interpreted as someone being enrolled in a school. 

\ex.\label{ex:baser-school}
\gll Okul-a kaydedil-en müdür-ün yeğen-i bahçe-de oyna-yor-du. \\
     school-\gl{dat} enroll-\gl{pass.rel} principal-\gl{gen} nephew-\gl{poss}.3\gl{sg} garden-\gl{loc} play-\gl{prog}-\gl{pst} \\
\glt `The nephew of the principal who was enrolled in the school was playing in the garden.'
\hfill (\citealt[251]{baser2018syntactic})

In our design, by contrast, the critical manipulation differs in a theoretically important way: \emph{both} candidate attachment analyses are constructed to remain pragmatically and semantically \emph{viable}. Concretely, for each item, the relative clause (RC) can plausibly attach to either NP1 or NP2 without yielding a categorical anomaly or selectional clash in contrast to \ref{ex:baser-school}. Thus, the manipulation does not instantiate a binary contrast between a \emph{possible} parse and an \emph{implausible} parse. Rather, it keeps the underlying syntactic configuration constant and manipulates a \emph{graded} asymmetry in \emph{world-knowledge plausibility}: local event knowledge is used to make one attachment interpretation \emph{more plausible} than its competitor, while preserving the acceptability of the alternative interpretation. Methodologically, this design supports a more fine-grained test of attachment processing. Because neither parse can be dismissed outright, any preference for HA vs.\ LA is less likely to reflect simple anomaly rejection and more likely to reflect sensitivity to subtle plausibility weighting during online interpretation. In turn, this allows us to ask whether human comprehenders and LLMs exploit probabilistic world knowledge as a soft constraint in attachment resolution, rather than relying only on cases in which one candidate parse is pragmatically untenable.

\ex.\label{ex:wk-plausibility}
\gll Gazeteci-ler-den kaç-an manken-in koruma-sı güçlü-ydü.\\
     reporter-\textsc{pl}-\textsc{abl} run.away-\textsc{rel} model-\textsc{gen} bodyguard-\textsc{poss} strong-\textsc{pst}\\
\glt `The bodyguard of the model, who ran away from reporters, was very strong.'

\noindent
In \ref{ex:wk-plausibility}, both \emph{manken} `model' and \emph{koruma} `bodyguard' are structurally viable RC hosts, but world knowledge makes the \emph{model} a much more plausible agent of \emph{kaç-} `run away' than the bodyguard, thereby creating a targeted plausibility cue that directly bears on attachment choice. Lexical frequency was controlled by consulting corpus-based frequency estimates for critical nouns and verbs and avoiding systematic imbalances across conditions; animacy was matched as well.

The critical materials comprised 40 ambiguous Turkish prenominal RC items (20 \textsc{High-WK}, 20 \textsc{Low-WK}). Across conditions, the syntactic configuration was held constant so that \emph{both} potential nominal hosts (N1 vs.\ N2) were structurally available; the manipulation targeted a narrower, graded \emph{world-knowledge plausibility} cue rather than broad pragmatic \emph{possibility}. Concretely, items were constructed so that attachment to one host would be strongly favored by local event plausibility (``who plausibly does what''), while attachment to the alternative host remained syntactically licensed and broadly possible. We validated this plausibility contrast with independent norming ratings.

\paragraph{Task and procedure.}
The experiment used a speeded forced-choice comprehension paradigm. On each trial, participants read an ambiguous RC sentence and then answered a \emph{who}-question designed to force an attachment decision by selecting the intended RC host (N1 vs.\ N2). The model evaluation does not reproduce this task literally. Instead, it operationalizes attachment preference by comparing the relative probability of two short continuations that unambiguously instantiate the HA and LA readings. We therefore treat the human task and the model evaluation as parallel decision settings over the same underlying ambiguity, rather than as identical online measures of processing.

\paragraph{LLM experiments: log-probability scoring.}
In \S\ref{sec:models}, we evaluate whether LLMs show plausibility-driven shifts over the same items by comparing the relative probability of matched HA and LA continuations. This does not constitute the same task as the human forced-choice experiment; rather, it provides a parallel model-side proxy for attachment preference under the same item manipulation.

\ex.\label{ex:mean-logprob}
\begin{equation*}
s(\mathbf{y}\mid x)
\;=\;
\frac{1}{k}\sum_{t=1}^{k}
\log P\!\left(y_t \mid x, y_1, \ldots, y_{t-1}\right).
\end{equation*}

We then define the log-probability preference:

\ex.\label{ex:delta}
\begin{equation*}
\Delta
\;=\;
s(\mathbf{y}^{\textsc{HA}}\mid x)
\;-\;
s(\mathbf{y}^{\textsc{LA}}\mid x),
\end{equation*}

and take the model’s predicted attachment to be \textsc{HA} iff $\Delta>0$ (otherwise \textsc{LA}). This yields a categorical HA/LA outcome per item, which we analyze analogously to the human choices via logistic regression of HA on WK condition.

\begin{figure}[t]
\centering
\begin{tikzpicture}[
  font=\footnotesize,
  node distance=3mm,
  arr/.style={-Latex, line width=0.6pt, draw=black!55},
  boxwide/.style={
    draw=black!45,
    rounded corners,
    fill=blue!3,
    align=left,
    inner sep=4pt,
    text width=0.92\columnwidth
  },
  boxhalf/.style={
    draw=black!45,
    rounded corners,
    fill=blue!2,
    align=left,
    inner sep=4pt,
    text width=0.44\columnwidth
  },
  boxfinal/.style={
    draw=black!55,
    rounded corners,
    fill=blue!4,
    align=left,
    inner sep=4pt,
    text width=0.92\columnwidth
  }
]

\node[boxwide] (stim) {\textbf{(1) Stimuli}\\
Ambiguous Turkish prenominal RCs with two candidate hosts (N1/HA vs.\ N2/LA).\\
Manipulation: world-knowledge plausibility favors HA vs.\ LA.};

\node[boxwide, below=of stim] (norm) {\textbf{(2) Norming}\\
Independent plausibility ratings validate the intended HA/LA bias.};

\node[boxhalf, below=8mm of norm, xshift=-0.24\columnwidth] (human)
{\textbf{(3a) Human task (Ibex)}\\
Sentence $\rightarrow$ \emph{who}-question\\
Choose N1 (HA) vs.\ N2 (LA).\\
Outcome: choice.};

\node[boxhalf, below=8mm of norm, xshift=+0.24\columnwidth] (llm)
{\textbf{(3b) LLM task}\\
Matched forced-choice completions.\\
Compute $\log p(\text{HA})$ vs.\ $\log p(\text{LA})$.\\
Outcome: model preference.};

\node[boxfinal, below=8mm of $(human.south)!0.5!(llm.south)$] (cmp)
{\textbf{(4) Comparison}\\
Quantify plausibility-driven shift in HA vs.\ LA: humans vs.\ LLMs.};

\draw[arr] (stim) -- (norm);

\draw[arr] (norm.south west) |- (human.north);
\draw[arr] (norm.south east) |- (llm.north);

\draw[arr] (human.south) |- (cmp.north west);
\draw[arr] (llm.south) |- (cmp.north east);

\end{tikzpicture}
\caption{Procedure overview. After constructing syntactically matched Turkish RC attachment ambiguities and validating plausibility via norming, we evaluate (a) human attachment choices in a speeded forced-choice task and (b) LLM attachment preferences via log-probability scoring over matched HA/LA continuations.}
\label{fig:procedure}
\end{figure}
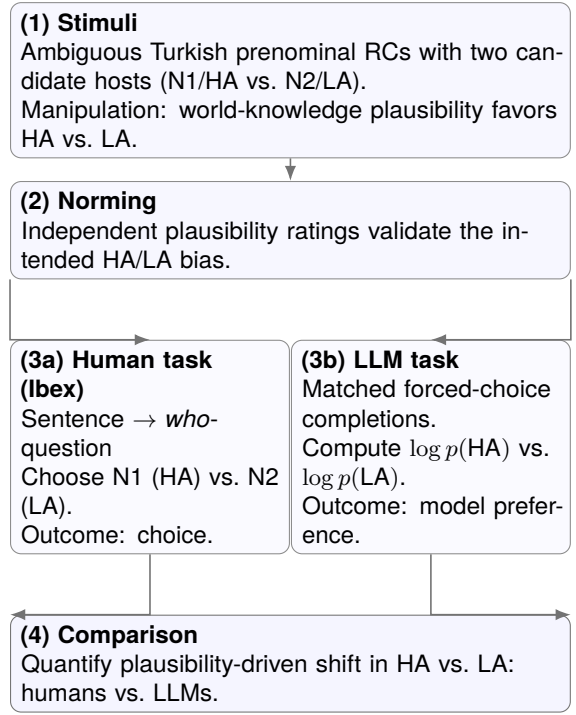

\subsection{Models}
\label{sec:models}

We evaluate attachment preferences in autoregressive transformer language models by scoring the two matched HA/LA continuations described above using token-level conditional log-probabilities. We include two Turkish-specific checkpoints and one strong multilingual checkpoint:

Ytu-ce-cosmos/turkish-gpt2. A Turkish GPT-2 style decoder-only language model used as a lightweight Turkish baseline.~\citep{kesgin2024introducingcosmosgptmonolingualtraining,ytucecosmosHF2024turkishgpt2}

Duxx/DeepSeek-R1-Distill-Qwen-1.5B-Turkish. A Turkish-adapted reasoning-oriented model obtained by fine-tuning a distilled DeepSeek-R1 checkpoint (Qwen-1.5B backbone) on Turkish reasoning data. We use it as a compact model that may better reflect structured inference than a plain GPT-2 baseline.~\citep{deepseekai2025deepseekr1,duxxHF2025deepseekr1distillqwen15bturkish}

Qwen/Qwen3-30B-A3B-Instruct-2507. A multilingual mixture-of-experts instruction-tuned model (30.5B total parameters, with 3.3B activated per token). We include this model to test whether a higher-capacity multilingual system shows more human-like sensitivity to world-knowledge plausibility in Turkish attachment.~\citep{yang2025qwen3technicalreport,qwenHF2025qwen3_30b_a3b_instruct_2507}

Recent Turkish benchmarking places Qwen3-30B-Instruct among the stronger publicly reported multilingual models for Turkish. TurkBench evaluates 27 open-source models on 8{,}151 instances across 21 subtasks, and its leaderboard sorts systems by the \emph{Avg} metric, that is, the overall average across task results; on this measure, Qwen3-30B-Instruct scores 73.4, compared with 78.6 for the top-ranked gpt-oss-120b \citep{toraman2026turkbench}. Because several higher-scoring systems were not tractable in our setup, we use Qwen3-30B-Instruct as a strong multilingual comparison point.

\section{Human Experiment Results}

\begin{figure}[t]
\centering

\begin{minipage}{\columnwidth}
\centering
\begin{tikzpicture}[x=1cm,y=4cm, font=\footnotesize]
\useasboundingbox (-0.8,-0.23) rectangle (3.1,1.10);

\def\lowha{0.263}
\def\highha{0.652}

\draw[->, line width=0.4pt] (-0.6,0) -- (-0.6,1.05) node[above] {HA rate};
\draw[->, line width=0.4pt] (-0.6,0) -- (3.0,0);

\foreach \y/\lab in {0/0\%,0.25/25\%,0.5/50\%,0.75/75\%,1/100\%}{
  \draw[line width=0.4pt] (-0.62,\y) -- (-0.58,\y);
  \node[anchor=east] at (-0.64,\y) {\lab};
  \draw[gray!25, line width=0.3pt] (-0.58,\y) -- (2.8,\y);
}

\fill[green!70]  (0.2,0) rectangle (1.2,\lowha);
\fill[purple!70] (1.8,0) rectangle (2.8,\highha);

\node[above] at (0.7,\lowha)  {26.3\%};
\node[above] at (2.3,\highha) {65.2\%};

\node[below] at (0.7,0) {Low-WK};
\node[below] at (2.3,0) {High-WK};
\end{tikzpicture}

\par\smallskip
(a) HA by WK condition
\end{minipage}

\vspace{4mm} 

\begin{minipage}{\columnwidth}
\centering
\begin{tikzpicture}[x=1cm,y=4cm, font=\footnotesize]
\useasboundingbox (-0.8,-0.23) rectangle (3.1,1.10);

\def\lowla{0.737}
\def\highla{0.348}

\draw[->, line width=0.4pt] (-0.6,0) -- (-0.6,1.05) node[above] {LA rate};
\draw[->, line width=0.4pt] (-0.6,0) -- (3.0,0);

\foreach \y/\lab in {0/0\%,0.25/25\%,0.5/50\%,0.75/75\%,1/100\%}{
  \draw[line width=0.4pt] (-0.62,\y) -- (-0.58,\y);
  \node[anchor=east] at (-0.64,\y) {\lab};
  \draw[gray!25, line width=0.3pt] (-0.58,\y) -- (2.8,\y);
}

\fill[green!70]  (0.2,0) rectangle (1.2,\lowla);
\fill[purple!70] (1.8,0) rectangle (2.8,\highla);

\node[above] at (0.7,\lowla)  {73.7\%};
\node[above] at (2.3,\highla) {34.8\%};

\node[below] at (0.7,0) {Low-WK};
\node[below] at (2.3,0) {High-WK};
\end{tikzpicture}

\par\smallskip
(b) LA by WK condition
\end{minipage}

\caption{Human attachment rates by world-knowledge (WK) condition. Panel (a) shows HA rates; panel (b) shows the complementary LA rates (100--HA).}
\label{fig:human-by-wk}
\end{figure}

Figure~\ref{fig:human-by-wk} summarizes attachment choices by world-knowledge
(WK) condition. In Low-WK contexts, where plausibility favors low attachment,
participants selected high attachment (HA) on 26.3\% of trials (panel a),
corresponding to a low-attachment (LA) rate of 73.7\% (panel b). In High-WK
contexts, where plausibility favors HA, HA increased to 65.2\% and LA decreased
to 34.8\%. This yields a 38.9 percentage-point shift in HA (65.2--26.3) and a
mirror-image shift in LA (34.8--73.7), showing that graded plausibility cues
robustly reweight attachment preferences even though both parses remain
pragmatically possible in our materials.

To test whether WK reliably modulated attachment, we fit a logistic regression
predicting HA (vs.\ LA) from WK condition. The effect was large and highly
reliable ($\beta_{\textsc{WK}}=1.65\pm0.11$, $z=14.75$, $p<10^{-50}$),
corresponding to an odds ratio of $e^{1.65}\approx 5.2$ (a 5.2$\times$ increase
in the odds of HA in High-WK relative to Low-WK). Collapsing across conditions,
the overall HA rate was 45.7\%, consistent with a modest baseline tendency
toward low attachment.

Robustness checks converged with the regression results. A contingency analysis
confirmed a strong association between WK condition and attachment choice
($\chi^{2}_{(1)}=229.1$, $p=1.7\times10^{-53}$). Moreover, across items, the
strength of the plausibility manipulation as measured by independent norming
ratings strongly tracked HA rates (Spearman $\rho=0.85$, $p=2\times10^{-6}$),
indicating that graded world-knowledge plausibility reliably shapes attachment
decisions in Turkish.

\subsection{Model results}
\label{sec:model-results}

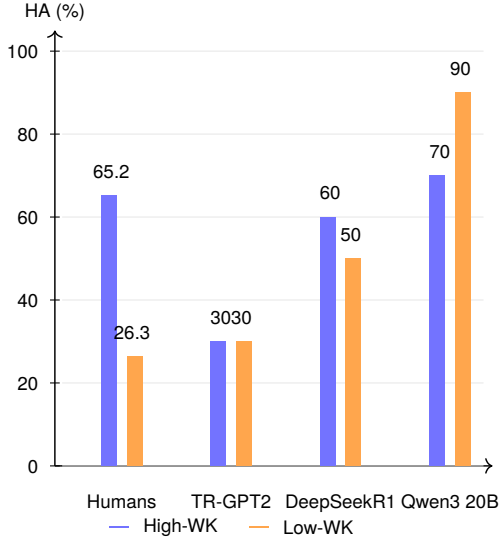
\begin{figure}[t]
\centering
\begin{tikzpicture}[x=0.80cm,y=0.055cm, font=\scriptsize]

\def\ymax{105}
\def\xmax{7.2}
\useasboundingbox (-0.6,-18) rectangle (\xmax+0.3,\ymax+2);

\draw[->, line width=0.55pt] (0,0) -- (0,\ymax);
\draw[->, line width=0.55pt] (0,0) -- (\xmax,0);
\node[anchor=south] at (0,\ymax) {HA (\%)};

\foreach \y in {0,20,40,60,80,100} {
  \draw (-0.08,\y) -- (0.08,\y);
  \node[anchor=east] at (-0.12,\y) {\y};
  \draw[gray!20, line width=0.25pt] (0.12,\y) -- (\xmax-0.15,\y);
}

\def\xH{1.1}
\def\xG{2.9}
\def\xD{4.7}
\def\xQ{6.5}

\def\bw{0.26}
\def\gap{0.08}
\def\laboff{2.0}

\newcommand{\twobars}[3]{%
  \fill[blue!55]   (#1-\bw-\gap,0) rectangle (#1-\gap, #2);
  \fill[orange!70] (#1+\gap,0) rectangle (#1+\bw+\gap, #3);
  \node[anchor=south] at (#1-\bw/2-\gap/2, #2+\laboff) {#2};
  \node[anchor=south] at (#1+\bw/2+\gap/2, #3+\laboff) {#3};
}

\twobars{\xH}{65.2}{26.3}  
\twobars{\xG}{30}{30}      
\twobars{\xD}{60}{50}      
\twobars{\xQ}{70}{90}      

\node[anchor=north, align=center] at (\xH,-4.6) {Humans};
\node[anchor=north, align=center] at (\xG,-4.6) {TR-GPT2};
\node[anchor=north, align=center] at (\xD,-4.6) {DeepSeekR1};
\node[anchor=north, align=center] at (\xQ,-4.6) {Qwen3 20B};

\begin{scope}[shift={(0.9,-15.0)}]
  \fill[blue!55] (0,0) rectangle (0.32,0.32);
  \node[anchor=west] at (0.40,0.16) {High-WK};
  \fill[orange!70] (2.3,0) rectangle (2.62,0.32);
  \node[anchor=west] at (2.70,0.16) {Low-WK};
\end{scope}

\end{tikzpicture}
\caption{HA rates (\%) by WK condition (High-WK vs.\ Low-WK) for humans and models.}
\label{fig:model-ha-only}
\end{figure}

We evaluated model attachment preferences using the continuation-based log-probability comparison described in \S\ref{sec:models}. For each item, we scored a high-attachment continuation and a low-attachment continuation and predicted HA when the HA continuation had higher mean per-token conditional log-probability (see \ref{ex:mean-logprob}--\ref{ex:delta}). Figure~\ref{fig:model-ha-only} summarizes HA choice rates by condition (blue: High-WK; orange: Low-WK). 

The human pattern in Figure~\ref{fig:model-ha-only} shows a large, correctly directed plausibility effect: HA increases from 26.3\% in Low-WK to 65.2\% in High-WK. In contrast, the models show substantially weaker and less human-like modulation. Turkish GPT-2 exhibits a stable LA tendency: HA remains at 30\% in both High-WK and Low-WK, with the Low-WK-only breakdown showing 70\% LA and 30\% HA in contexts intended to favor LA. DeepSeek-R1-Distill-Qwen-1.5B-Turkish shows only a small shift (HA 60\% in High-WK vs.\ 50\% in Low-WK), and its Low-WK-only breakdown is essentially balanced (50\% HA vs.\ 50\% LA). Qwen3 shows a different failure mode on the 20-item pilot (10 High-WK + 10 Low-WK): a strong overall HA bias and a reversed direction relative to the manipulation (HA 70\% in High-WK but 90\% in Low-WK), which is also visible in Figure~\ref{fig:model-ha-only} as only 10\% LA in Low-WK items.

To summarize world-knowledge sensitivity across systems, we report the difference in HA rate across conditions:
$\Delta\mathrm{HA} = \mathrm{HA}_{\textsc{High-WK}} - \mathrm{HA}_{\textsc{Low-WK}}$.

\ex.\label{ex:delta-ha-summary}
\begin{align*}
\Delta\mathrm{HA}_{\text{Humans}} \;&=\; +38.9\text{ pp} \quad (65.2 - 26.3) \\
\Delta\mathrm{HA}_{\text{TR-GPT2}} \;&=\; 0.0\text{ pp} \quad (30 - 30) \\
\Delta\mathrm{HA}_{\text{DeepSeek}} \;&=\; +10.0\text{ pp} \quad (60 - 50) \\
\Delta\mathrm{HA}_{\text{Qwen3-20}} \;&=\; -20.0\text{ pp} \quad (70 - 90)
\end{align*}

We tested condition effects on the categorical HA/LA outcomes using Fisher exact tests on the 2$\times$2 contingency table (condition $\times$ attachment). Turkish GPT-2 shows no condition effect (6/20 HA in High-WK vs.\ 6/20 HA in Low-WK; $p=1.0$). DeepSeek shows no reliable condition effect at this sample size (12/20 HA in High-WK vs.\ 10/20 HA in Low-WK; $p=0.751$). Qwen3 also does not show a reliable categorical difference under Fisher’s test on the 10+10 pilot (7/10 HA in High-WK vs.\ 9/10 HA in Low-WK; $p=0.582$), despite the visibly reversed direction in Figure~\ref{fig:model-ha-only}.

For Qwen3, we also analyzed the continuous log-probability margin to quantify
how strongly the model prefers one attachment over the other. For each item
$i$, we define the HA--LA margin as:

\ex.\label{ex:qwen3-margin}
\begin{equation*}
m_i \;=\; s(\mathbf{y}^{\textsc{HA}}\mid x_i)\;-\;s(\mathbf{y}^{\textsc{LA}}\mid x_i).
\end{equation*}

Larger $m_i$ indicates a stronger preference for HA. Comparing margins across
conditions reveals a statistically detectable effect, but in the wrong
direction: Low-WK items are more HA-favoring than High-WK items. Concretely,
the mean difference satisfies:

\ex.\label{ex:qwen3-margin-diff}
\begin{equation*}
\mathbb{E}\!\left[m_{\textsc{Low-WK}} - m_{\textsc{High-WK}}\right] \;\approx\; 5.74
\;\text{nats},
\end{equation*}

with a bootstrap 95\% interval approximately $[1.15, 9.96]$. A Welch t-test
gives $p=0.0266$ and a Mann--Whitney U test gives $p=0.0211$. Thus, even when
the condition effect is detectable at the margin level, it reflects increased
HA preference in Low-WK contexts, consistent with the reversed pattern in
Figure~\ref{fig:model-ha-only}.

Overall, the tests indicate that the models do not reproduce the strong, correctly directed plausibility-driven shift observed in humans. Instead, Turkish GPT-2 shows a rigid LA bias with no modulation, DeepSeek shows weak and unreliable modulation at this sample size, and Qwen3 shows a strong HA bias with a reversed condition effect on the pilot set.

\section{Discussion}
\label{sec:discussion}

Our results reveal a sharp human--model dissociation in how graded world knowledge is used to resolve Turkish RC attachment ambiguity.
Humans show a large, correctly directed plausibility effect: HA rises from 26.3\% in Low-WK to 65.2\% in High-WK (a +38.9 percentage-point shift), with the complementary LA plot showing the mirror-image decrease.
By contrast, all tested LMs exhibit weak or qualitatively different behavior.
In the abstract baselines, Turkish GPT-2 shows a stable LA bias (HA 30\% in both conditions), while DeepSeek-R1-Distill-Qwen-1.5B-Turkish shows only a small shift (HA 60\% vs.\ 50\%).
Qwen3, despite being a much stronger general-purpose model, shows a strong overall HA bias and, on our 20-item subset, a reversed WK effect (HA 70\% in High-WK vs.\ 90\% in Low-WK), yielding negative WK sensitivity.

A natural interpretation is that LLMs can store substantial commonsense and factual knowledge, yet deploy it unreliably in incremental ambiguity resolution where syntactic structure and plausibility must be integrated under tight constraints. At the same time, the comparison should be interpreted at the level of attachment preference rather than as a direct process-level alignment. Human participants performed a speeded comprehension task with an explicit question, whereas models were evaluated through relative preference for matched continuations. The value of the comparison, therefore, lies in asking whether the same plausibility manipulation shifts interpretation in the same direction across humans and models, not in claiming that the two systems were probed with an identical online measure.

This aligns with recent cognitive-science arguments that linguistic proficiency and human-like conceptual reasoning can dissociate in LLMs, motivating careful use of targeted behavioral diagnostics rather than relying on broad benchmark success alone \citep{mahowald2024dissociating}.
In our setting, the plausibility manipulation is deliberately subtle: both parses remain pragmatically possible, so the task is not to reject an anomalous interpretation but to reweight two licensed structures using graded event knowledge. Humans do so robustly, but the models show much weaker and less consistent plausibility-driven shifts, suggesting that plausibility information is not being used as reliably as it is in human attachment resolution.

\paragraph{Reconciling with ``LLMs are good at commonsense'' results.}
At first glance, our findings may appear in tension with results showing strong LLM performance on commonsense and world-knowledge benchmarks, including large technical reports documenting broad benchmark strength for frontier models \citep{grattafiori2024llama3}, and recent multilingual physical-commonsense evaluations where state-of-the-art LLMs perform well in aggregate \citep{chang2025globalpiqa}.
There is also growing evidence that \emph{scaffolding} methods can substantially improve commonsense QA (e.g., guided knowledge generation) \citep{wei2024guidekg}.
We argue that these successes are compatible with our dissociation because they often probe \emph{knowledge access under explicit QA framing} (and sometimes with additional guidance), whereas Turkish RC attachment requires \emph{knowledge deployment in real-time structure building}.
In other words, passing a commonsense benchmark does not guarantee that the model will (i) retrieve the relevant event knowledge, (ii) align it with syntactic roles, and (iii) use it to \emph{reweight} two grammatically licensed parses when neither interpretation is outright anomalous.
Our paradigm targets exactly this ``commonsense-in-use'' requirement and shows that, relative to humans, current LLMs do not robustly integrate graded plausibility with syntactic attachment.
In that sense, our study supports the broader conclusion that LLM commonsense reasoning remains unreliable: not necessarily because the knowledge is absent, but because it is not consistently \emph{applied} when the task demands subtle, structurally constrained disambiguation.

Besides, our Turkish findings are consistent with a growing line of work arguing that attachment ambiguities remain a sensitive probe of LLM inductive biases.
Recent multilingual studies report that models often default to local attachment and show limited responsiveness to language-specific or structure-sensitive patterns, with strong effects of evaluation framing and prompt format \citep{luo2025multiwho}.
Work focusing specifically on semantic bias in RC attachment similarly suggests that LLMs can miss or inconsistently apply plausibility cues that humans use, with variability across models and setups \citep{scheinberg2025missingcues}.
Hence, these findings support a picture in which LLM behavior on ambiguity resolution is not well predicted by their general benchmark strength, but is shaped by a mixture of base-rate structural tendencies, data-driven lexical associations, and sensitivity to how the disambiguation question is posed.

\subsection{Why might Turkish be especially challenging?}
One possibility is that Turkish prenominal RCs place heavier demands on anticipating structure before the head noun, and the relevant plausibility information is distributed across morphosyntax and lexical roles \citep{ozge2015incremental}.
If training data provides weaker or noisier supervision for these configurations (relative to high-resource Indo-European patterns), models may fall back on simpler heuristics that do not incorporate graded plausibility in a human-like way.
Another possibility is that instruction-tuned models tend to produce overly peaked choice distributions even when multiple continuations are viable, which can interact with our aligned scoring protocol and yield a rigid bias (as in Qwen3) \citep{zhang2024forcingdiffuse}.
Distinguishing these explanations requires systematically varying continuation format, question framing, and the amount and position of plausibility-supporting context.

 Furthermore, tokenization matters especially for morphologically rich and agglutinative languages such as Turkish, where many cues relevant to interpretation are carried by suffix chains and where subword tokenizers can fragment roots and affixes in ways that are only weakly aligned with linguistic units. In Turkish, prior work shows that tokenizer choice can substantially affect model behavior and downstream performance, and that simple morphological-level tokenization does not automatically yield better language modeling than subword approaches \citep{toraman2023impacttokenization,bayram2025tokenizationc,bayram2025tokstandards,bayram2025tokensmeaning}. Complementary evidence on Turkish LLMs likewise reports measurable differences attributable to tokenization granularity, reinforcing that tokenizer design can alter what a model treats as an atomic cue and how efficiently it represents Turkish wordforms \citep{kaya2024effecttokenizationgranularity}.

\section{Conclusion}
\label{sec:conclusion}

We presented a controlled, cross-population test of how graded world-knowledge plausibility shapes relative-clause attachment in Turkish prenominal RC ambiguities. Using normed materials in which \emph{both} parses remained pragmatically possible, we showed that native Turkish speakers robustly integrated event plausibility in attachment resolution: HA increased by 38.9 percentage points from Low-WK to High-WK. In contrast, the three autoregressive LLMs evaluated with a continuation-based log-probability comparison showed substantially weaker, less consistent, and in one case reversed sensitivity to the same plausibility manipulation. Overall, the model results suggest that graded plausibility did not guide attachment preferences as reliably as it did in human judgments.

These findings are consistent with the possibility that, in the tested models, plausibility information is not incorporated as reliably when attachment resolution requires tightly constrained integration of world knowledge with hierarchical structure. More broadly, the results suggest that strong overall benchmark performance does not by itself guarantee human-like cue integration in psycholinguistic tasks, and that attachment ambiguities under subtle plausibility manipulations can provide a useful diagnostic for comparing human and model behavior in language understanding.

\section*{Limitations}
\label{sec:limitations}

First, the stimulus set is modest, which limits power to detect small plausibility effects and reduces confidence in fine-grained cross-model comparisons. Second, the model suite is necessarily selective and excludes the very strongest Turkish-capable systems, so our conclusions are restricted to the tested checkpoints. Finally, we compare categorical human choices to model preferences rather than directly matching incremental time-course measures; future work may add stronger online proxies such as region-wise surprisal profiles to more closely link models to human processing. Finally, future work should also investigate whether corpus evidence can help clarify baseline attachment tendencies in Turkish and better contextualize the experimental results.

\section*{Acknowledgments}

I thank Chris Kennedy for valuable discussions in his \emph{Linguistics and LLMs} class. I am also grateful to the Human Sentence Processing Conference 2026 at MIT for feedback and discussion, and to Muharrem Taha Aydın and Mustafa Baki Varol for their help with the design, discussion of the results, and feedback. Finally, I thank the anonymous reviewers for their thoughtful comments and suggestions.

\section{Bibliographical References}\label{sec:reference}

\bibliographystyle{lrec2026-natbib}
\bibliography{lrec2026-example}

@inproceedings{hale2001,
  title     = {A Probabilistic Earley Parser as a Psycholinguistic Model},
  author    = {Hale, John},
  booktitle = {Second Meeting of the North American Chapter of the Association for Computational Linguistics on Language Technologies},
  year      = {2001},
  doi       = {10.3115/1073336.1073357},
  url       = {https://aclanthology.org/N01-1021/}
}

@article{levy2008,
  title   = {Expectation-based syntactic comprehension},
  author  = {Levy, Roger},
  journal = {Cognition},
  year    = {2008},
  volume  = {106},
  number  = {3},
  pages   = {1126--1177},
  doi     = {10.1016/j.cognition.2007.05.006}
}

@inproceedings{bayram2025tokenizationc,
  author    = {Bayram, M. A. and Fincan, A. A. and G{\"u}m{\"u}{\c s}, A. S. and Karaka{\c s}, S. and Diri, B. and Y{\i}ld{\i}r{\i}m, S.},
  title     = {Tokenization Standards and Evaluation in Natural Language Processing: A Comparative Analysis of Large Language Models on Turkish},
  booktitle = {2025 33rd Signal Processing and Communications Applications Conference (SIU)},
  pages     = {1--4},
  year      = {2025},
  month     = jun,
  publisher = {IEEE},
  url       = {https://ieeexplore.ieee.org/abstract/document/11112220/}
}

@inproceedings{arehalli-etal-2022-syntactic,
  title     = {Syntactic Surprisal From Neural Models Predicts, But Underestimates, Human Processing Difficulty From Syntactic Ambiguities},
  author    = {Arehalli, Suhas and Dillon, Brian and Linzen, Tal},
  booktitle = {Proceedings of the 26th Conference on Computational Natural Language Learning (CoNLL)},
  year      = {2022},
  pages     = {301--313},
  publisher = {Association for Computational Linguistics},
  doi       = {10.18653/v1/2022.conll-1.20},
  url       = {https://aclanthology.org/2022.conll-1.20/}
}

@inproceedings{lake2018generalization,
  title     = {Generalization without Systematicity: On the Compositional Skills of Sequence-to-Sequence Recurrent Networks},
  author    = {Lake, Brenden M. and Baroni, Marco},
  booktitle = {Proceedings of the 35th International Conference on Machine Learning (ICML)},
  year      = {2018},
  url       = {https://proceedings.mlr.press/v80/lake18a.html}
}

@inproceedings{ruis2020gscan,
  title     = {A Benchmark for Systematic Generalization in Grounded Language Understanding},
  author    = {Ruis, Laura and Andreas, Jacob and Baroni, Marco and Bouchacourt, Diane and Lake, Brenden M.},
  booktitle = {Advances in Neural Information Processing Systems (NeurIPS)},
  year      = {2020},
  url       = {https://arxiv.org/abs/2003.05161}
}

@inproceedings{mccoy-etal-2019-right,
  title     = {Right for the Wrong Reasons: Diagnosing Syntactic Heuristics in Natural Language Inference},
  author    = {McCoy, R. Thomas and Pavlick, Ellie and Linzen, Tal},
  booktitle = {Proceedings of the 57th Annual Meeting of the Association for Computational Linguistics},
  pages     = {3428--3448},
  year      = {2019},
  address   = {Florence, Italy},
  publisher = {Association for Computational Linguistics},
  doi       = {10.18653/v1/P19-1334},
  url       = {https://aclanthology.org/P19-1334/}
}

@misc{zehrschwarz2018penncontroller,
  author       = {Zehr, Jeremy and Schwarz, Florian},
  title        = {PennController for Internet Based Experiments (IBEX)},
  year         = {2018},
  publisher    = {OSF},
  doi          = {10.17605/OSF.IO/MD832}
}

@inproceedings{yang-etal-2024-exploring,
  title     = {Exploring Compositional Generalization of Large Language Models},
  author    = {Yang, Haoran and Lu, Hongyuan and Lam, Wai and Cai, Deng},
  booktitle = {Proceedings of the 2024 Conference of the North American Chapter of the Association for Computational Linguistics: Human Language Technologies (Volume 4: Student Research Workshop)},
  pages     = {16--24},
  year      = {2024},
  month     = jun,
  address   = {Mexico City, Mexico},
  publisher = {Association for Computational Linguistics},
  doi       = {10.18653/v1/2024.naacl-srw.3},
  url       = {https://aclanthology.org/2024.naacl-srw.3/}
}

@phdthesis{baser2018syntactic,
  author = {Başer, Zeynep},
  title = {Syntactic Priming of Relative Clause Attachment in Monolingual Turkish Speakers and Turkish Learners of English},
  school = {Middle East Technical University},
  year = {2018},
  url = {https://open.metu.edu.tr/handle/11511/27252}
}

@misc{openai2024gpt4technicalreport,
      title={GPT-4 Technical Report}, 
      author={OpenAI and Josh Achiam and Steven Adler and Sandhini Agarwal and Lama Ahmad and Ilge Akkaya and Florencia Leoni Aleman and Diogo Almeida and Janko Altenschmidt and Sam Altman and Shyamal Anadkat and Red Avila and Igor Babuschkin and Suchir Balaji and Valerie Balcom and Paul Baltescu and Haiming Bao and Mohammad Bavarian and Jeff Belgum and Irwan Bello and Jake Berdine and Gabriel Bernadett-Shapiro and Christopher Berner and Lenny Bogdonoff and Oleg Boiko and Madelaine Boyd and Anna-Luisa Brakman and Greg Brockman and Tim Brooks and Miles Brundage and Kevin Button and Trevor Cai and Rosie Campbell and Andrew Cann and Brittany Carey and Chelsea Carlson and Rory Carmichael and Brooke Chan and Che Chang and Fotis Chantzis and Derek Chen and Sully Chen and Ruby Chen and Jason Chen and Mark Chen and Ben Chess and Chester Cho and Casey Chu and Hyung Won Chung and Dave Cummings and Jeremiah Currier and Yunxing Dai and Cory Decareaux and Thomas Degry and Noah Deutsch and Damien Deville and Arka Dhar and David Dohan and Steve Dowling and Sheila Dunning and Adrien Ecoffet and Atty Eleti and Tyna Eloundou and David Farhi and Liam Fedus and Niko Felix and Simón Posada Fishman and Juston Forte and Isabella Fulford and Leo Gao and Elie Georges and Christian Gibson and Vik Goel and Tarun Gogineni and Gabriel Goh and Rapha Gontijo-Lopes and Jonathan Gordon and Morgan Grafstein and Scott Gray and Ryan Greene and Joshua Gross and Shixiang Shane Gu and Yufei Guo and Chris Hallacy and Jesse Han and Jeff Harris and Yuchen He and Mike Heaton and Johannes Heidecke and Chris Hesse and Alan Hickey and Wade Hickey and Peter Hoeschele and Brandon Houghton and Kenny Hsu and Shengli Hu and Xin Hu and Joost Huizinga and Shantanu Jain and Shawn Jain and Joanne Jang and Angela Jiang and Roger Jiang and Haozhun Jin and Denny Jin and Shino Jomoto and Billie Jonn and Heewoo Jun and Tomer Kaftan and Łukasz Kaiser and Ali Kamali and Ingmar Kanitscheider and Nitish Shirish Keskar and Tabarak Khan and Logan Kilpatrick and Jong Wook Kim and Christina Kim and Yongjik Kim and Jan Hendrik Kirchner and Jamie Kiros and Matt Knight and Daniel Kokotajlo and Łukasz Kondraciuk and Andrew Kondrich and Aris Konstantinidis and Kyle Kosic and Gretchen Krueger and Vishal Kuo and Michael Lampe and Ikai Lan and Teddy Lee and Jan Leike and Jade Leung and Daniel Levy and Chak Ming Li and Rachel Lim and Molly Lin and Stephanie Lin and Mateusz Litwin and Theresa Lopez and Ryan Lowe and Patricia Lue and Anna Makanju and Kim Malfacini and Sam Manning and Todor Markov and Yaniv Markovski and Bianca Martin and Katie Mayer and Andrew Mayne and Bob McGrew and Scott Mayer McKinney and Christine McLeavey and Paul McMillan and Jake McNeil and David Medina and Aalok Mehta and Jacob Menick and Luke Metz and Andrey Mishchenko and Pamela Mishkin and Vinnie Monaco and Evan Morikawa and Daniel Mossing and Tong Mu and Mira Murati and Oleg Murk and David Mély and Ashvin Nair and Reiichiro Nakano and Rajeev Nayak and Arvind Neelakantan and Richard Ngo and Hyeonwoo Noh and Long Ouyang and Cullen O'Keefe and Jakub Pachocki and Alex Paino and Joe Palermo and Ashley Pantuliano and Giambattista Parascandolo and Joel Parish and Emy Parparita and Alex Passos and Mikhail Pavlov and Andrew Peng and Adam Perelman and Filipe de Avila Belbute Peres and Michael Petrov and Henrique Ponde de Oliveira Pinto and Michael and Pokorny and Michelle Pokrass and Vitchyr H. Pong and Tolly Powell and Alethea Power and Boris Power and Elizabeth Proehl and Raul Puri and Alec Radford and Jack Rae and Aditya Ramesh and Cameron Raymond and Francis Real and Kendra Rimbach and Carl Ross and Bob Rotsted and Henri Roussez and Nick Ryder and Mario Saltarelli and Ted Sanders and Shibani Santurkar and Girish Sastry and Heather Schmidt and David Schnurr and John Schulman and Daniel Selsam and Kyla Sheppard and Toki Sherbakov and Jessica Shieh and Sarah Shoker and Pranav Shyam and Szymon Sidor and Eric Sigler and Maddie Simens and Jordan Sitkin and Katarina Slama and Ian Sohl and Benjamin Sokolowsky and Yang Song and Natalie Staudacher and Felipe Petroski Such and Natalie Summers and Ilya Sutskever and Jie Tang and Nikolas Tezak and Madeleine B. Thompson and Phil Tillet and Amin Tootoonchian and Elizabeth Tseng and Preston Tuggle and Nick Turley and Jerry Tworek and Juan Felipe Cerón Uribe and Andrea Vallone and Arun Vijayvergiya and Chelsea Voss and Carroll Wainwright and Justin Jay Wang and Alvin Wang and Ben Wang and Jonathan Ward and Jason Wei and CJ Weinmann and Akila Welihinda and Peter Welinder and Jiayi Weng and Lilian Weng and Matt Wiethoff and Dave Willner and Clemens Winter and Samuel Wolrich and Hannah Wong and Lauren Workman and Sherwin Wu and Jeff Wu and Michael Wu and Kai Xiao and Tao Xu and Sarah Yoo and Kevin Yu and Qiming Yuan and Wojciech Zaremba and Rowan Zellers and Chong Zhang and Marvin Zhang and Shengjia Zhao and Tianhao Zheng and Juntang Zhuang and William Zhuk and Barret Zoph},
      year={2024},
      eprint={2303.08774},
      archivePrefix={arXiv},
      primaryClass={cs.CL},
      url={https://arxiv.org/abs/2303.08774}, 
}

@misc{finkelstein2026translategemmatechnicalreport,
      title={TranslateGemma Technical Report}, 
      author={Mara Finkelstein and Isaac Caswell and Tobias Domhan and Jan-Thorsten Peter and Juraj Juraska and Parker Riley and Daniel Deutsch and Geza Kovacs and Cole Dilanni and Colin Cherry and Eleftheria Briakou and Elizabeth Nielsen and Jiaming Luo and Kat Black and Ryan Mullins and Sweta Agrawal and Wenda Xu and Erin Kats and Stephane Jaskiewicz and Markus Freitag and David Vilar},
      year={2026},
      eprint={2601.09012},
      archivePrefix={arXiv},
      primaryClass={cs.CL},
      url={https://arxiv.org/abs/2601.09012}, 
}

@misc{singh2025openaigpt5card,
      title={OpenAI GPT-5 System Card}, 
      author={Aaditya Singh and Adam Fry and Adam Perelman and Adam Tart and Adi Ganesh and Ahmed El-Kishky and Aidan McLaughlin and Aiden Low and AJ Ostrow and Akhila Ananthram and Akshay Nathan and Alan Luo and Alec Helyar and Aleksander Madry and Aleksandr Efremov and Aleksandra Spyra and Alex Baker-Whitcomb and Alex Beutel and Alex Karpenko and Alex Makelov and Alex Neitz and Alex Wei and Alexandra Barr and Alexandre Kirchmeyer and Alexey Ivanov and Alexi Christakis and Alistair Gillespie and Allison Tam and Ally Bennett and Alvin Wan and Alyssa Huang and Amy McDonald Sandjideh and Amy Yang and Ananya Kumar and Andre Saraiva and Andrea Vallone and Andrei Gheorghe and Andres Garcia Garcia and Andrew Braunstein and Andrew Liu and Andrew Schmidt and Andrey Mereskin and Andrey Mishchenko and Andy Applebaum and Andy Rogerson and Ann Rajan and Annie Wei and Anoop Kotha and Anubha Srivastava and Anushree Agrawal and Arun Vijayvergiya and Ashley Tyra and Ashvin Nair and Avi Nayak and Ben Eggers and Bessie Ji and Beth Hoover and Bill Chen and Blair Chen and Boaz Barak and Borys Minaiev and Botao Hao and Bowen Baker and Brad Lightcap and Brandon McKinzie and Brandon Wang and Brendan Quinn and Brian Fioca and Brian Hsu and Brian Yang and Brian Yu and Brian Zhang and Brittany Brenner and Callie Riggins Zetino and Cameron Raymond and Camillo Lugaresi and Carolina Paz and Cary Hudson and Cedric Whitney and Chak Li and Charles Chen and Charlotte Cole and Chelsea Voss and Chen Ding and Chen Shen and Chengdu Huang and Chris Colby and Chris Hallacy and Chris Koch and Chris Lu and Christina Kaplan and Christina Kim and CJ Minott-Henriques and Cliff Frey and Cody Yu and Coley Czarnecki and Colin Reid and Colin Wei and Cory Decareaux and Cristina Scheau and Cyril Zhang and Cyrus Forbes and Da Tang and Dakota Goldberg and Dan Roberts and Dana Palmie and Daniel Kappler and Daniel Levine and Daniel Wright and Dave Leo and David Lin and David Robinson and Declan Grabb and Derek Chen and Derek Lim and Derek Salama and Dibya Bhattacharjee and Dimitris Tsipras and Dinghua Li and Dingli Yu and DJ Strouse and Drew Williams and Dylan Hunn and Ed Bayes and Edwin Arbus and Ekin Akyurek and Elaine Ya Le and Elana Widmann and Eli Yani and Elizabeth Proehl and Enis Sert and Enoch Cheung and Eri Schwartz and Eric Han and Eric Jiang and Eric Mitchell and Eric Sigler and Eric Wallace and Erik Ritter and Erin Kavanaugh and Evan Mays and Evgenii Nikishin and Fangyuan Li and Felipe Petroski Such and Filipe de Avila Belbute Peres and Filippo Raso and Florent Bekerman and Foivos Tsimpourlas and Fotis Chantzis and Francis Song and Francis Zhang and Gaby Raila and Garrett McGrath and Gary Briggs and Gary Yang and Giambattista Parascandolo and Gildas Chabot and Grace Kim and Grace Zhao and Gregory Valiant and Guillaume Leclerc and Hadi Salman and Hanson Wang and Hao Sheng and Haoming Jiang and Haoyu Wang and Haozhun Jin and Harshit Sikchi and Heather Schmidt and Henry Aspegren and Honglin Chen and Huida Qiu and Hunter Lightman and Ian Covert and Ian Kivlichan and Ian Silber and Ian Sohl and Ibrahim Hammoud and Ignasi Clavera and Ikai Lan and Ilge Akkaya and Ilya Kostrikov and Irina Kofman and Isak Etinger and Ishaan Singal and Jackie Hehir and Jacob Huh and Jacqueline Pan and Jake Wilczynski and Jakub Pachocki and James Lee and James Quinn and Jamie Kiros and Janvi Kalra and Jasmyn Samaroo and Jason Wang and Jason Wolfe and Jay Chen and Jay Wang and Jean Harb and Jeffrey Han and Jeffrey Wang and Jennifer Zhao and Jeremy Chen and Jerene Yang and Jerry Tworek and Jesse Chand and Jessica Landon and Jessica Liang and Ji Lin and Jiancheng Liu and Jianfeng Wang and Jie Tang and Jihan Yin and Joanne Jang and Joel Morris and Joey Flynn and Johannes Ferstad and Johannes Heidecke and John Fishbein and John Hallman and Jonah Grant and Jonathan Chien and Jonathan Gordon and Jongsoo Park and Jordan Liss and Jos Kraaijeveld and Joseph Guay and Joseph Mo and Josh Lawson and Josh McGrath and Joshua Vendrow and Joy Jiao and Julian Lee and Julie Steele and Julie Wang and Junhua Mao and Kai Chen and Kai Hayashi and Kai Xiao and Kamyar Salahi and Kan Wu and Karan Sekhri and Karan Sharma and Karan Singhal and Karen Li and Kenny Nguyen and Keren Gu-Lemberg and Kevin King and Kevin Liu and Kevin Stone and Kevin Yu and Kristen Ying and Kristian Georgiev and Kristie Lim and Kushal Tirumala and Kyle Miller and Lama Ahmad and Larry Lv and Laura Clare and Laurance Fauconnet and Lauren Itow and Lauren Yang and Laurentia Romaniuk and Leah Anise and Lee Byron and Leher Pathak and Leon Maksin and Leyan Lo and Leyton Ho and Li Jing and Liang Wu and Liang Xiong and Lien Mamitsuka and Lin Yang and Lindsay McCallum and Lindsey Held and Liz Bourgeois and Logan Engstrom and Lorenz Kuhn and Louis Feuvrier and Lu Zhang and Lucas Switzer and Lukas Kondraciuk and Lukasz Kaiser and Manas Joglekar and Mandeep Singh and Mandip Shah and Manuka Stratta and Marcus Williams and Mark Chen and Mark Sun and Marselus Cayton and Martin Li and Marvin Zhang and Marwan Aljubeh and Matt Nichols and Matthew Haines and Max Schwarzer and Mayank Gupta and Meghan Shah and Melody Huang and Meng Dong and Mengqing Wang and Mia Glaese and Micah Carroll and Michael Lampe and Michael Malek and Michael Sharman and Michael Zhang and Michele Wang and Michelle Pokrass and Mihai Florian and Mikhail Pavlov and Miles Wang and Ming Chen and Mingxuan Wang and Minnia Feng and Mo Bavarian and Molly Lin and Moose Abdool and Mostafa Rohaninejad and Nacho Soto and Natalie Staudacher and Natan LaFontaine and Nathan Marwell and Nelson Liu and Nick Preston and Nick Turley and Nicklas Ansman and Nicole Blades and Nikil Pancha and Nikita Mikhaylin and Niko Felix and Nikunj Handa and Nishant Rai and Nitish Keskar and Noam Brown and Ofir Nachum and Oleg Boiko and Oleg Murk and Olivia Watkins and Oona Gleeson and Pamela Mishkin and Patryk Lesiewicz and Paul Baltescu and Pavel Belov and Peter Zhokhov and Philip Pronin and Phillip Guo and Phoebe Thacker and Qi Liu and Qiming Yuan and Qinghua Liu and Rachel Dias and Rachel Puckett and Rahul Arora and Ravi Teja Mullapudi and Raz Gaon and Reah Miyara and Rennie Song and Rishabh Aggarwal and RJ Marsan and Robel Yemiru and Robert Xiong and Rohan Kshirsagar and Rohan Nuttall and Roman Tsiupa and Ronen Eldan and Rose Wang and Roshan James and Roy Ziv and Rui Shu and Ruslan Nigmatullin and Saachi Jain and Saam Talaie and Sam Altman and Sam Arnesen and Sam Toizer and Sam Toyer and Samuel Miserendino and Sandhini Agarwal and Sarah Yoo and Savannah Heon and Scott Ethersmith and Sean Grove and Sean Taylor and Sebastien Bubeck and Sever Banesiu and Shaokyi Amdo and Shengjia Zhao and Sherwin Wu and Shibani Santurkar and Shiyu Zhao and Shraman Ray Chaudhuri and Shreyas Krishnaswamy and Shuaiqi and Xia and Shuyang Cheng and Shyamal Anadkat and Simón Posada Fishman and Simon Tobin and Siyuan Fu and Somay Jain and Song Mei and Sonya Egoian and Spencer Kim and Spug Golden and SQ Mah and Steph Lin and Stephen Imm and Steve Sharpe and Steve Yadlowsky and Sulman Choudhry and Sungwon Eum and Suvansh Sanjeev and Tabarak Khan and Tal Stramer and Tao Wang and Tao Xin and Tarun Gogineni and Taya Christianson and Ted Sanders and Tejal Patwardhan and Thomas Degry and Thomas Shadwell and Tianfu Fu and Tianshi Gao and Timur Garipov and Tina Sriskandarajah and Toki Sherbakov and Tomer Kaftan and Tomo Hiratsuka and Tongzhou Wang and Tony Song and Tony Zhao and Troy Peterson and Val Kharitonov and Victoria Chernova and Vineet Kosaraju and Vishal Kuo and Vitchyr Pong and Vivek Verma and Vlad Petrov and Wanning Jiang and Weixing Zhang and Wenda Zhou and Wenlei Xie and Wenting Zhan and Wes McCabe and Will DePue and Will Ellsworth and Wulfie Bain and Wyatt Thompson and Xiangning Chen and Xiangyu Qi and Xin Xiang and Xinwei Shi and Yann Dubois and Yaodong Yu and Yara Khakbaz and Yifan Wu and Yilei Qian and Yin Tat Lee and Yinbo Chen and Yizhen Zhang and Yizhong Xiong and Yonglong Tian and Young Cha and Yu Bai and Yu Yang and Yuan Yuan and Yuanzhi Li and Yufeng Zhang and Yuguang Yang and Yujia Jin and Yun Jiang and Yunyun Wang and Yushi Wang and Yutian Liu and Zach Stubenvoll and Zehao Dou and Zheng Wu and Zhigang Wang},
      year={2025},
      eprint={2601.03267},
      archivePrefix={arXiv},
      primaryClass={cs.CL},
      url={https://arxiv.org/abs/2601.03267}, 
}

@misc{riviere2024gemma2,
      title={Gemma 2: Improving Open Language Models at a Practical Size}, 
      author={Gemma Team and Morgane Riviere and Shreya Pathak and Pier Giuseppe Sessa and Cassidy Hardin and Surya Bhupatiraju and Léonard Hussenot and Thomas Mesnard and Bobak Shahriari and Alexandre Ramé and Johan Ferret and Peter Liu and Pouya Tafti and Abe Friesen and Michelle Casbon and Sabela Ramos and Ravin Kumar and Charline Le Lan and Sammy Jerome and Anton Tsitsulin and Nino Vieillard and Piotr Stanczyk and Sertan Girgin and Nikola Momchev and Matt Hoffman and Shantanu Thakoor and Jean-Bastien Grill and Behnam Neyshabur and Olivier Bachem and Alanna Walton and Aliaksei Severyn and Alicia Parrish and Aliya Ahmad and Allen Hutchison and Alvin Abdagic and Amanda Carl and Amy Shen and Andy Brock and Andy Coenen and Anthony Laforge and Antonia Paterson and Ben Bastian and Bilal Piot and Bo Wu and Brandon Royal and Charlie Chen and Chintu Kumar and Chris Perry and Chris Welty and Christopher A. Choquette-Choo and Danila Sinopalnikov and David Weinberger and Dimple Vijaykumar and Dominika Rogozińska and Dustin Herbison and Elisa Bandy and Emma Wang and Eric Noland and Erica Moreira and Evan Senter and Evgenii Eltyshev and Francesco Visin and Gabriel Rasskin and Gary Wei and Glenn Cameron and Gus Martins and Hadi Hashemi and Hanna Klimczak-Plucińska and Harleen Batra and Harsh Dhand and Ivan Nardini and Jacinda Mein and Jack Zhou and James Svensson and Jeff Stanway and Jetha Chan and Jin Peng Zhou and Joana Carrasqueira and Joana Iljazi and Jocelyn Becker and Joe Fernandez and Joost van Amersfoort and Josh Gordon and Josh Lipschultz and Josh Newlan and Ju-yeong Ji and Kareem Mohamed and Kartikeya Badola and Kat Black and Katie Millican and Keelin McDonell and Kelvin Nguyen and Kiranbir Sodhia and Kish Greene and Lars Lowe Sjoesund and Lauren Usui and Laurent Sifre and Lena Heuermann and Leticia Lago and Lilly McNealus and Livio Baldini Soares and Logan Kilpatrick and Lucas Dixon and Luciano Martins and Machel Reid and Manvinder Singh and Mark Iverson and Martin Görner and Mat Velloso and Mateo Wirth and Matt Davidow and Matt Miller and Matthew Rahtz and Matthew Watson and Meg Risdal and Mehran Kazemi and Michael Moynihan and Ming Zhang and Minsuk Kahng and Minwoo Park and Mofi Rahman and Mohit Khatwani and Natalie Dao and Nenshad Bardoliwalla and Nesh Devanathan and Neta Dumai and Nilay Chauhan and Oscar Wahltinez and Pankil Botarda and Parker Barnes and Paul Barham and Paul Michel and Pengchong Jin and Petko Georgiev and Phil Culliton and Pradeep Kuppala and Ramona Comanescu and Ramona Merhej and Reena Jana and Reza Ardeshir Rokni and Rishabh Agarwal and Ryan Mullins and Samaneh Saadat and Sara Mc Carthy and Sarah Cogan and Sarah Perrin and Sébastien M. R. Arnold and Sebastian Krause and Shengyang Dai and Shruti Garg and Shruti Sheth and Sue Ronstrom and Susan Chan and Timothy Jordan and Ting Yu and Tom Eccles and Tom Hennigan and Tomas Kocisky and Tulsee Doshi and Vihan Jain and Vikas Yadav and Vilobh Meshram and Vishal Dharmadhikari and Warren Barkley and Wei Wei and Wenming Ye and Woohyun Han and Woosuk Kwon and Xiang Xu and Zhe Shen and Zhitao Gong and Zichuan Wei and Victor Cotruta and Phoebe Kirk and Anand Rao and Minh Giang and Ludovic Peran and Tris Warkentin and Eli Collins and Joelle Barral and Zoubin Ghahramani and Raia Hadsell and D. Sculley and Jeanine Banks and Anca Dragan and Slav Petrov and Oriol Vinyals and Jeff Dean and Demis Hassabis and Koray Kavukcuoglu and Clement Farabet and Elena Buchatskaya and Sebastian Borgeaud and Noah Fiedel and Armand Joulin and Kathleen Kenealy and Robert Dadashi and Alek Andreev},
      year={2024},
      eprint={2408.00118},
      archivePrefix={arXiv},
      primaryClass={cs.CL},
      url={https://arxiv.org/abs/2408.00118}, 

    }

@misc{anil2023palm2,
      title={PaLM 2 Technical Report}, 
      author={Rohan Anil and Andrew M. Dai and Orhan Firat and Melvin Johnson and Dmitry Lepikhin and Alexandre Passos and Siamak Shakeri and Emanuel Taropa and Paige Bailey and Zhifeng Chen and Eric Chu and Jonathan H. Clark and Laurent El Shafey and Yanping Huang and Kathy Meier-Hellstern and Gaurav Mishra and Erica Moreira and Mark Omernick and Kevin Robinson and Sebastian Ruder and Yi Tay and Kefan Xiao and Yuanzhong Xu and Yujing Zhang and Gustavo Hernandez Abrego and Junwhan Ahn and Jacob Austin and Paul Barham and Jan Botha and James Bradbury and Siddhartha Brahma and Kevin Brooks and Michele Catasta and Yong Cheng and Colin Cherry and Christopher A. Choquette-Choo and Aakanksha Chowdhery and Clément Crepy and Shachi Dave and Mostafa Dehghani and Sunipa Dev and Jacob Devlin and Mark Díaz and Nan Du and Ethan Dyer and Vlad Feinberg and Fangxiaoyu Feng and Vlad Fienber and Markus Freitag and Xavier Garcia and Sebastian Gehrmann and Lucas Gonzalez and Guy Gur-Ari and Steven Hand and Hadi Hashemi and Le Hou and Joshua Howland and Andrea Hu and Jeffrey Hui and Jeremy Hurwitz and Michael Isard and Abe Ittycheriah and Matthew Jagielski and Wenhao Jia and Kathleen Kenealy and Maxim Krikun and Sneha Kudugunta and Chang Lan and Katherine Lee and Benjamin Lee and Eric Li and Music Li and Wei Li and YaGuang Li and Jian Li and Hyeontaek Lim and Hanzhao Lin and Zhongtao Liu and Frederick Liu and Marcello Maggioni and Aroma Mahendru and Joshua Maynez and Vedant Misra and Maysam Moussalem and Zachary Nado and John Nham and Eric Ni and Andrew Nystrom and Alicia Parrish and Marie Pellat and Martin Polacek and Alex Polozov and Reiner Pope and Siyuan Qiao and Emily Reif and Bryan Richter and Parker Riley and Alex Castro Ros and Aurko Roy and Brennan Saeta and Rajkumar Samuel and Renee Shelby and Ambrose Slone and Daniel Smilkov and David R. So and Daniel Sohn and Simon Tokumine and Dasha Valter and Vijay Vasudevan and Kiran Vodrahalli and Xuezhi Wang and Pidong Wang and Zirui Wang and Tao Wang and John Wieting and Yuhuai Wu and Kelvin Xu and Yunhan Xu and Linting Xue and Pengcheng Yin and Jiahui Yu and Qiao Zhang and Steven Zheng and Ce Zheng and Weikang Zhou and Denny Zhou and Slav Petrov and Yonghui Wu},
      year={2023},
      eprint={2305.10403},
      archivePrefix={arXiv},
      primaryClass={cs.CL},
      url={https://arxiv.org/abs/2305.10403}, 
}

@misc{li2025appleintelligencefoundationlanguage,
      title={Apple Intelligence Foundation Language Models: Tech Report 2025}, 
      author={Ethan Li and Anders Boesen Lindbo Larsen and Chen Zhang and Xiyou Zhou and Jun Qin and Dian Ang Yap and Narendran Raghavan and Xuankai Chang and Margit Bowler and Eray Yildiz and John Peebles and Hannah Gillis Coleman and Matteo Ronchi and Peter Gray and Keen You and Anthony Spalvieri-Kruse and Ruoming Pang and Reed Li and Yuli Yang and Emad Soroush and Zhiyun Lu and Crystal Xiao and Rong Situ and Jordan Huffaker and David Griffiths and Zaid Ahmed and Peng Zhang and Daniel Parilla and Asaf Liberman and Jennifer Mallalieu and Parsa Mazaheri and Qibin Chen and Manjot Bilkhu and Aonan Zhang and Eric Wang and Dave Nelson and Michael FitzMaurice and Thomas Voice and Jeremy Liu and Josh Shaffer and Shiwen Zhao and Prasanth Yadla and Farzin Rasteh and Pengsheng Guo and Arsalan Farooq and Jeremy Snow and Stephen Murphy and Tao Lei and Minsik Cho and George Horrell and Sam Dodge and Lindsay Hislop and Sumeet Singh and Alex Dombrowski and Aiswarya Raghavan and Sasha Sirovica and Mandana Saebi and Faye Lao and Max Lam and TJ Lu and Zhaoyang Xu and Karanjeet Singh and Marc Kirchner and David Mizrahi and Rajat Arora and Haotian Zhang and Henry Mason and Lawrence Zhou and Yi Hua and Ankur Jain and Felix Bai and Joseph Astrauskas and Floris Weers and Josh Gardner and Mira Chiang and Yi Zhang and Pulkit Agrawal and Tony Sun and Quentin Keunebroek and Matthew Hopkins and Bugu Wu and Tao Jia and Chen Chen and Xingyu Zhou and Nanzhu Wang and Peng Liu and Ruixuan Hou and Rene Rauch and Yuan Gao and Afshin Dehghan and Jonathan Janke and Zirui Wang and Cha Chen and Xiaoyi Ren and Feng Nan and Josh Elman and Dong Yin and Yusuf Goren and Jeff Lai and Yiran Fei and Syd Evans and Muyang Yu and Guoli Yin and Yi Qin and Erin Feldman and Isha Garg and Aparna Rajamani and Karla Vega and Walker Cheng and TJ Collins and Hans Han and Raul Rea Menacho and Simon Yeung and Sophy Lee and Phani Mutyala and Ying-Chang Cheng and Zhe Gan and Sprite Chu and Justin Lazarow and Alessandro Pappalardo and Federico Scozzafava and Jing Lu and Erik Daxberger and Laurent Duchesne and Jen Liu and David Güera and Stefano Ligas and Mary Beth Kery and Brent Ramerth and Ciro Sannino and Marcin Eichner and Haoshuo Huang and Rui Qian and Moritz Schwarzer-Becker and David Riazati and Mingfei Gao and Bailin Wang and Jack Cackler and Yang Lu and Ransen Niu and John Dennison and Guillaume Klein and Jeffrey Bigham and Deepak Gopinath and Navid Shiee and Darren Botten and Guillaume Tartavel and Alex Guillen Garcia and Sam Xu and Victoria MönchJuan Haladjian and Zi-Yi Dou and Matthias Paulik and Adolfo Lopez Mendez and Zhen Li and Hong-You Chen and Chao Jia and Dhaval Doshi and Zhengdong Zhang and Raunak Manjani and Aaron Franklin and Zhile Ren and David Chen and Artsiom Peshko and Nandhitha Raghuram and Hans Hao and Jiulong Shan and Kavya Nerella and Ramsey Tantawi and Vivek Kumar and Saiwen Wang and Brycen Wershing and Bhuwan Dhingra and Dhruti Shah and Ob Adaranijo and Xin Zheng and Tait Madsen and Hadas Kotek and Chang Liu and Yin Xia and Hanli Li and Suma Jayaram and Yanchao Sun and Ahmed Fakhry and Vasileios Saveris and Dustin Withers and Yanghao Li and Alp Aygar and Andres Romero Mier Y Teran and Kaiwei Huang and Mark Lee and Xiujun Li and Yuhong Li and Tyler Johnson and Jay Tang and Joseph Yitan Cheng and Futang Peng and Andrew Walkingshaw and Lucas Guibert and Abhishek Sharma and Cheng Shen and Piotr Maj and Yasutaka Tanaka and You-Cyuan Jhang and Vivian Ma and Tommi Vehvilainen and Kelvin Zou and Jeff Nichols and Matthew Lei and David Qiu and Yihao Qian and Gokul Santhanam and Wentao Wu and Yena Han and Dominik Moritz and Haijing Fu and Mingze Xu and Vivek Rathod and Jian Liu and Louis D'hauwe and Qin Ba and Haitian Sun and Haoran Yan and Philipp Dufter and Anh Nguyen and Yihao Feng and Emma Wang and Keyu He and Rahul Nair and Sanskruti Shah and Jiarui Lu and Patrick Sonnenberg and Jeremy Warner and Yuanzhi Li and Bowen Pan and Ziyi Zhong and Joe Zhou and Sam Davarnia and Olli Saarikivi and Irina Belousova and Rachel Burger and Shang-Chen Wu and Di Feng and Bas Straathof and James Chou and Yuanyang Zhang and Marco Zuliani and Eduardo Jimenez and Abhishek Sundararajan and Xianzhi Du and Chang Lan and Nilesh Shahdadpuri and Peter Grasch and Sergiu Sima and Josh Newnham and Varsha Paidi and Jianyu Wang and Kaelen Haag and Alex Braunstein and Daniele Molinari and Richard Wei and Brenda Yang and Nicholas Lusskin and Joanna Arreaza-Taylor and Meng Cao and Nicholas Seidl and Simon Wang and Jiaming Hu and Yiping Ma and Mengyu Li and Kieran Liu and Hang Su and Sachin Ravi and Chong Wang and Xin Wang and Kevin Smith and Haoxuan You and Binazir Karimzadeh and Rui Li and Jinhao Lei and Wei Fang and Alec Doane and Sam Wiseman and Ismael Fernandez and Jane Li and Andrew Hansen and Javier Movellan and Christopher Neubauer and Hanzhi Zhou and Chris Chaney and Nazir Kamaldin and Valentin Wolf and Fernando Bermúdez-Medina and Joris Pelemans and Peter Fu and Howard Xing and Xiang Kong and Wayne Shan and Gabriel Jacoby-Cooper and Dongcai Shen and Tom Gunter and Guillaume Seguin and Fangping Shi and Shiyu Li and Yang Xu and Areeba Kamal and Dan Masi and Saptarshi Guha and Qi Zhu and Jenna Thibodeau and Changyuan Zhang and Rebecca Callahan and Charles Maalouf and Wilson Tsao and Boyue Li and Qingqing Cao and Naomy Sabo and Cheng Leong and Yi Wang and Anupama Mann Anupama and Colorado Reed and Kenneth Jung and Zhifeng Chen and Mohana Prasad Sathya Moorthy and Yifei He and Erik Hornberger and Devi Krishna and Senyu Tong and Michael and Lee and David Haldimann and Yang Zhao and Bowen Zhang and Chang Gao and Chris Bartels and Sushma Rao and Nathalie Tran and Simon Lehnerer and Co Giang and Patrick Dong and Junting Pan and Biyao Wang and Dongxu Li and Mehrdad Farajtabar and Dongseong Hwang and Grace Duanmu and Eshan Verma and Sujeeth Reddy and Qi Shan and Hongbin Gao and Nan Du and Pragnya Sridhar and Forrest Huang and Yingbo Wang and Nikhil Bhendawade and Diane Zhu and Sai Aitharaju and Fred Hohman and Lauren Gardiner and Chung-Cheng Chiu and Yinfei Yang and Alper Kokmen and Frank Chu and Ke Ye and Kaan Elgin and Oron Levy and John Park and Donald Zhang and Eldon Schoop and Nina Wenzel and Michael Booker and Hyunjik Kim and Chinguun Erdenebileg and Nan Dun and Eric Liang Yang and Priyal Chhatrapati and Vishaal Mahtani and Haiming Gang and Kohen Chia and Deepa Seshadri and Donghan Yu and Yan Meng and Kelsey Peterson and Zhen Yang and Yongqiang Wang and Carina Peng and Doug Kang and Anuva Agarwal and Albert Antony and Juan Lao Tebar and Albin Madappally Jose and Regan Poston and Andy De Wang and Gerard Casamayor and Elmira Amirloo and Violet Yao and Wojciech Kryscinski and Kun Duan and Lezhi L},
      year={2025},
      eprint={2507.13575},
      archivePrefix={arXiv},
      primaryClass={cs.LG},
      url={https://arxiv.org/abs/2507.13575}, 
}

@inproceedings{futrell-etal-2019-neural,
  title     = "Neural language models as psycholinguistic subjects: Representations of syntactic state",
  author    = "Futrell, Richard  and Wilcox, Ethan  and Morita, Takashi  and Qian, Peng  and Ballesteros, Miguel  and Levy, Roger",
  booktitle = "Proceedings of the 2019 Conference of the North American Chapter of the Association for Computational Linguistics: Human Language Technologies, Volume 1 (Long and Short Papers)",
  month     = jun,
  year      = "2019",
  address   = "Minneapolis, Minnesota",
  publisher = "Association for Computational Linguistics",
  url       = "https://aclanthology.org/N19-1004/",
  doi       = "10.18653/v1/N19-1004",
  pages     = "32--42"
}

@inproceedings{joshi-etal-2020-state,
  title     = "The State and Fate of Linguistic Diversity and Inclusion in the {NLP} World",
  author    = "Joshi, Pratik  and Santy, Sebastin  and Budhiraja, Amar  and Bali, Kalika  and Choudhury, Monojit",
  booktitle = "Proceedings of the 58th Annual Meeting of the Association for Computational Linguistics",
  month     = jul,
  year      = "2020",
  address   = "Online",
  publisher = "Association for Computational Linguistics",
  url       = "https://aclanthology.org/2020.acl-main.560/",
  doi       = "10.18653/v1/2020.acl-main.560",
  pages     = "6282--6293"
}

@inproceedings{hollenstein-etal-2021-multilingual,
  title     = "Multilingual Language Models Predict Human Reading Behavior",
  author    = "Hollenstein, Nora  and Pirovano, Federico  and Zhang, Ce  and J{\"a}ger, Lena  and Beinborn, Lisa",
  booktitle = "Proceedings of the 2021 Conference of the North American Chapter of the Association for Computational Linguistics: Human Language Technologies",
  month     = jun,
  year      = "2021",
  address   = "Online",
  publisher = "Association for Computational Linguistics",
  url       = "https://aclanthology.org/2021.naacl-main.10/",
  doi       = "10.18653/v1/2021.naacl-main.10",
  pages     = "106--123"
}

@inproceedings{oh-schuler-2023-transformer,
  title     = "Transformer-Based Language Model Surprisal Predicts Human Reading Times Best with About Two Billion Training Tokens",
  author    = "Oh, Byung-Doh  and Schuler, William",
  booktitle = "Findings of the Association for Computational Linguistics: EMNLP 2023",
  month     = dec,
  year      = "2023",
  address   = "Singapore",
  publisher = "Association for Computational Linguistics",
  url       = "https://aclanthology.org/2023.findings-emnlp.128/",
  doi       = "10.18653/v1/2023.findings-emnlp.128",
  pages     = "1915--1921"
}

@inproceedings{alves-2025-benchmarking,
  title     = "Benchmarking Language Model Surprisal for Eye-Tracking Predictions in {B}razilian {P}ortuguese",
  author    = "Alves, Diego",
  booktitle = "Proceedings of the First International Workshop on Gaze Data and Natural Language Processing",
  month     = sep,
  year      = "2025",
  address   = "Varna, Bulgaria",
  publisher = "INCOMA Ltd., Shoumen, BULGARIA",
  url       = "https://aclanthology.org/2025.gaze4nlp-1.2/",
  pages     = "7--17"
}

@article{boeve-bogaerts-2025-dutch,
  title     = "A systematic evaluation of {D}utch large language models’ surprisal estimates in sentence, paragraph and book reading",
  author    = "Boeve, Sam and Bogaerts, Louisa",
  journal   = "Behavior Research Methods",
  year      = "2025",
  volume    = "57",
  pages     = "266",
  month     = aug,
  doi       = "10.3758/s13428-025-02774-4",
  url       = "https://link.springer.com/article/10.3758/s13428-025-02774-4",
  note      = "Published 18 Aug 2025; article number 266"
}

@inproceedings{conneau-etal-2020-unsupervised,
  title     = {Unsupervised Cross-lingual Representation Learning at Scale},
  author    = {Conneau, Alexis and Khandelwal, Kartikay and Goyal, Naman and
               Chaudhary, Vishrav and Wenzek, Guillaume and Guzm{\'a}n, Francisco and
               Grave, Edouard and Ott, Myle and Zettlemoyer, Luke and Stoyanov, Veselin},
  booktitle = {Proceedings of the 58th Annual Meeting of the Association for Computational Linguistics},
  month     = jul,
  year      = {2020},
  address   = {Online},
  publisher = {Association for Computational Linguistics},
  pages     = {8440--8451},
  doi       = {10.18653/v1/2020.acl-main.747},
  url       = {https://aclanthology.org/2020.acl-main.747/}
}

@article{spivey1993context,
  title   = {Context effects in syntactic ambiguity resolution: Discourse and semantic influences in parsing reduced relative clauses},
  author  = {Spivey-Knowlton, Michael J. and Trueswell, John C. and Tanenhaus, Michael K.},
  journal = {Canadian Journal of Experimental Psychology / Revue canadienne de psychologie exp{\'e}rimentale},
  year    = {1993},
  volume  = {47},
  number  = {2},
  pages   = {276--309},
  doi     = {10.1037/h0078826},
  url     = {https://pubmed.ncbi.nlm.nih.gov/8364532/},
  note    = {PsycNet record: https://psycnet.apa.org/record/1994-04330-001}
}

@article{frazierfodor1978sausagemachine,
  title   = {The Sausage Machine: A New Two-Stage Parsing Model},
  author  = {Frazier, Lyn and Fodor, Janet Dean},
  journal = {Cognition},
  year    = {1978},
  volume  = {6},
  number  = {4},
  pages   = {291--325},
  doi     = {10.1016/0010-0277(78)90002-1},
  url     = {https://www.sciencedirect.com/science/article/pii/0010027778900021}
}

@article{kimball1973seven,
  title   = {Seven principles of surface structure parsing in natural language},
  author  = {Kimball, John},
  journal = {Cognition},
  year    = {1973},
  volume  = {2},
  number  = {1},
  pages   = {15--47},
  doi     = {10.1016/0010-0277(72)90028-5},
  url     = {https://www.sciencedirect.com/science/article/pii/0010027772900285}
}

@article{gibson1996recency,
  title   = {Recency preference in the human sentence processing mechanism},
  author  = {Gibson, Edward and Pearlmutter, Neal J. and Canseco-Gonzalez, Elizabeth and Hickok, Gregory},
  journal = {Cognition},
  year    = {1996},
  volume  = {59},
  number  = {1},
  pages   = {23--59},
  doi     = {10.1016/0010-0277(95)00687-7},
  url     = {https://www.sciencedirect.com/science/article/pii/0010027795006877}
}

@incollection{baccino2000lateclosure,
  title     = {Cross-Linguistic Studies of the Late Closure Strategy: French and Italian},
  author    = {Baccino, Thierry and De Vincenzi, Marica and Job, Remo},
  booktitle = {Cross-Linguistic Perspectives on Language Processing},
  editor    = {De Vincenzi, Marica and Lombardo, Vincenzo},
  series    = {Studies in Theoretical Psycholinguistics},
  volume    = {25},
  publisher = {Springer},
  address   = {Dordrecht},
  year      = {2000},
  pages     = {89--118},
  doi       = {10.1007/978-94-011-3949-6_4},
  url       = {https://link.springer.com/chapter/10.1007/978-94-011-3949-6_4}
}

@book{fernandez2003bilingual,
  title     = {Bilingual Sentence Processing: Relative Clause Attachment in English and Spanish},
  author    = {Fern{\'a}ndez, Eva M.},
  series    = {Language Acquisition and Language Disorders},
  volume    = {29},
  publisher = {John Benjamins},
  address   = {Amsterdam/Philadelphia},
  year      = {2003},
  doi       = {10.1075/lald.29},
  url       = {https://www.jbe-platform.com/content/books/9789027296788}
}

@article{altmannsteedman1988context,
  title   = {Interaction with context during human sentence processing},
  author  = {Altmann, Gerry and Steedman, Mark},
  journal = {Cognition},
  year    = {1988},
  volume  = {30},
  number  = {3},
  pages   = {191--238},
  doi     = {10.1016/0010-0277(88)90020-0}
}

@article{mcrae1998thematicfit,
  title   = {Modeling the Influence of Thematic Fit (and Other Constraints) in On-line Sentence Comprehension},
  author  = {McRae, Ken and Spivey-Knowlton, Michael J. and Tanenhaus, Michael K.},
  journal = {Journal of Memory and Language},
  year    = {1998},
  volume  = {38},
  number  = {3},
  pages   = {283--312},
  doi     = {10.1006/jmla.1997.2543}
}

@article{kirkici2004turkishrc,
  title   = {The processing of relative clause attachment ambiguities in Turkish},
  author  = {K{\i}rk{\i}c{\i}, Bilal},
  journal = {Turkic Languages},
  year    = {2004},
  volume  = {8},
  note    = {Preprint PDF available online},
  url     = {https://turkoloji.cu.edu.tr/DILBILIM/bilal_kirkici_relative_clause.pdf}
}

@article{baser2020turkishrc,
  title   = {Is There a Particular {RC} Attachment Preference in {T}urkish? Negotiating the Effects of Semantic Factors},
  author  = {Ba{\c{s}}er, Zeynep and Hohenberger, Annette},
  journal = {Journal of Psycholinguistic Research},
  year    = {2020},
  volume  = {49},
  number  = {4},
  pages   = {511--539},
  doi     = {10.1007/s10936-020-09698-4}
}

@article{akal2021recency,
  title   = {Recency preference in ambiguous relative clause attachment in {Turkish}},
  author  = {Akal, Taylan},
  journal = {Journal of Language and Linguistic Studies},
  year    = {2021},
  volume  = {17},
  number  = {Special Issue 1},
  pages   = {139--159},
  doi     = {},
  url     = {https://files.eric.ed.gov/fulltext/EJ1284901.pdf}
}

@unpublished{logacev2022underspecification,
  title  = {Absence of Evidence for Underspecification in Prenominal Relative Clause Attachment},
  author = {Loga{\v{c}}ev, Pavel and {\"O}zgur Ayd{\i}n and Aylin M{\"u}ge Tuncer},
  year   = {2022},
  note   = {Manuscript (version dated January 8, 2022); includes Turkish eye-tracking and self-paced reading},
  url    = {https://avesis.ankara.edu.tr/yayin/5bb8d178-f2c9-4326-884c-3493e700df78/absence-of-evidence-for-underspecification-in-prenominal-relative-clause-attachment/document.pdf}
}

@inproceedings{davis2020rnncontext,
  title     = {Interaction with Context During Recurrent Neural Network Sentence Processing},
  author    = {Davis, Forrest and van Schijndel, Marten},
  booktitle = {Proceedings of the 42nd Annual Meeting of the Cognitive Science Society (CogSci 2020)},
  year      = {2020},
  note      = {Conference paper},
  url       = {https://cognitivesciencesociety.org/cogsci20/papers/0672/0672.pdf}
}

@article{ettinger2020bert,
  title   = {What {BERT} Is Not: Lessons from a New Suite of Psycholinguistic Diagnostics for Language Models},
  author  = {Ettinger, Allyson},
  journal = {Transactions of the Association for Computational Linguistics},
  year    = {2020},
  volume  = {8},
  pages   = {34--48},
  doi     = {10.1162/tacl_a_00298},
  url     = {https://aclanthology.org/2020.tacl-1.3/}
}

@inproceedings{de-varda-marelli-2022-effects,
  title     = {The Effects of Surprisal across Languages: Results from Native and Non-native Reading},
  author    = {de Varda, Andrea and Marelli, Marco},
  booktitle = {Findings of the Association for Computational Linguistics: AACL-IJCNLP 2022},
  year      = {2022},
  month     = nov,
  address   = {Online only},
  publisher = {Association for Computational Linguistics},
  pages     = {138--144},
  doi       = {10.18653/v1/2022.findings-aacl.13},
  url       = {https://aclanthology.org/2022.findings-aacl.13/}
}

@inproceedings{keles-deniz-2024-superficial,
  title     = {A Comparative Study with Human Data: Do {LLM}s Have Superficial Language Processing?},
  author    = {Kele{\c{s}}, Onur and Din{\c{c}}topal Deniz, Nazik},
  booktitle = {2024 32nd Signal Processing and Communications Applications Conference ({SIU})},
  year      = {2024},
  month     = may,
  address   = {Mersin, T{\"u}rkiye},
  publisher = {IEEE},
  note      = {IEEE Xplore document 10600807; proceedings held 15--18 May 2024 in Mersin, T{\"u}rkiye},
  url       = {https://ieeexplore.ieee.org/abstract/document/10600807/}
}

@misc{kesgin2024introducingcosmosgptmonolingualtraining,
      title={Introducing cosmosGPT: Monolingual Training for Turkish Language Models},
      author={H. Toprak Kesgin and M. Kaan Yuce and Eren Dogan and M. Egemen Uzun and Atahan Uz and H. Emre Seyrek and Ahmed Zeer and M. Fatih Amasyali},
      year={2024},
      eprint={2404.17336},
      archivePrefix={arXiv},
      primaryClass={cs.CL},
      url={https://arxiv.org/abs/2404.17336},
}

@misc{yang2025qwen3technicalreport,
      title={Qwen3 Technical Report},
      author={An Yang and Anfeng Li and Baosong Yang and Beichen Zhang and Binyuan Hui and Bo Zheng and Bowen Yu and Chang Gao and Chengen Huang and Chenxu Lv and Chujie Zheng and Dayiheng Liu and Fan Zhou and Fei Huang and Feng Hu and Hao Ge and Haoran Wei and Huan Lin and Jialong Tang and Jian Yang and Jianhong Tu and Jianwei Zhang and Jianxin Yang and Jiaxi Yang and Jing Zhou and Jingren Zhou and Junyang Lin and Kai Dang and Keqin Bao and Kexin Yang and Le Yu and Lianghao Deng and Mei Li and Mingfeng Xue and Mingze Li and Pei Zhang and Peng Wang and Qin Zhu and Rui Men and Ruize Gao and Shixuan Liu and Shuang Luo and Tianhao Li and Tianyi Tang and Wenbiao Yin and Xingzhang Ren and Xinyu Wang and Xinyu Zhang and Xuancheng Ren and Yang Fan and Yang Su and Yichang Zhang and Yinger Zhang and Yu Wan and Yuqiong Liu and Zekun Wang and Zeyu Cui and Zhenru Zhang and Zhipeng Zhou and Zihan Qiu},
      year={2025},
      eprint={2505.09388},
      archivePrefix={arXiv},
      primaryClass={cs.CL},
      url={https://arxiv.org/abs/2505.09388},
}

@misc{ytucecosmosHF2024turkishgpt2,
  author       = {{ytu-ce-cosmos}},
  title        = {{turkish-gpt2}},
  howpublished = {Hugging Face model card},
  year         = {2024},
  url          = {https://huggingface.co/ytu-ce-cosmos/turkish-gpt2},
  note         = {Accessed 2026-02-09}
}

@misc{duxxHF2025deepseekr1distillqwen15bturkish,
  author       = {{duxx}},
  title        = {{DeepSeek-R1-Distill-Qwen-1.5B-Turkish}},
  howpublished = {Hugging Face model card},
  year         = {2025},
  url          = {https://huggingface.co/duxx/DeepSeek-R1-Distill-Qwen-1.5B-Turkish},
  note         = {Accessed 2026-02-09}
}

@misc{qwenHF2025qwen3_30b_a3b_instruct_2507,
  author       = {{Qwen}},
  title        = {{Qwen3-30B-A3B-Instruct-2507}},
  howpublished = {Hugging Face model card},
  year         = {2025},
  url          = {https://huggingface.co/Qwen/Qwen3-30B-A3B-Instruct-2507},
  note         = {Accessed 2026-02-09}
}

@misc{toraman2026turkbench,
  title         = {TurkBench: A Benchmark for Evaluating Turkish Large Language Models},
  author        = {Toraman, {\c{C}}a{\u{g}}r{\i} and Sever, Ahmet Kaan and Cengiz, Ayse Aysu and Arslan, Elif Ecem and Sevin{\c{c}}, G{\"o}rkem and Birdal, Mete Mert and G{\"u}ldemir, Yusuf Faruk and Kanburo{\u{g}}lu, Ali Bu{\u{g}}ra and Feleko{\u{g}}lu, Sezen and G{\"u}rlek, Osman and Kantar, Sarp and {\c{S}}ahin K{\"u}t{\"u}k, Birsen and Tufan, B{\"u}{\c{s}}ra and Gen{\c{c}}, Elif and Co{\c{s}}kun, Serkan and Demir, Gupse Ekin and Aray{\i}c{\i}, Muhammed Emin and Dursun, Olgun and Gungor, Onur and {\"U}sk{\"u}darl{\i}, Susan and Topraksoy, Abdullah and Dar{\i}c{\i}, Esra},
  year          = {2026},
  eprint        = {2601.07020},
  archivePrefix = {arXiv},
  primaryClass  = {cs.CL},
  url           = {https://arxiv.org/abs/2601.07020}
}

\label{lr:ref}
\bibliographystylelanguageresource{lrec2026-natbib}
\bibliographylanguageresource{languageresource}

\end{document}